\documentclass[11pt,a4paper]{article}
\usepackage{times,latexsym}
\usepackage{url}
\usepackage[T1]{fontenc}
\usepackage{multirow}
\usepackage{graphicx}
\usepackage{amsmath}
\usepackage{booktabs}
\usepackage{xcolor, colortbl}
\usepackage{pifont}
\usepackage{comment}
\usepackage{subcaption}
\usepackage{tcolorbox}
\usepackage{soul}
\usepackage{xurl}
\usepackage{needspace}

  \usepackage[acceptedWithA]{tacl2021v1}

\usepackage{xspace,mfirstuc,tabulary}

\newif\iftaclinstructions
\taclinstructionsfalse 
\iftaclinstructions
\renewcommand{\confidential}{}
\renewcommand{\anonsubtext}{(No author info supplied here, for consistency with
TACL-submission anonymization requirements)}
\newcommand{\instr}
\fi

\iftaclpubformat 

\else

\fi

\title{What do vision-language models see in the context? Investigating multimodal in-context learning}

\author{
  Gabriel O. dos Santos\Thanks{Corresponding author.} \and
  Esther Colombini \and Sandra Avila   \\
  Instituto de Computação, Universidade Estadual de Campinas (UNICAMP), Campinas, Brazil \\
  \texttt{ \{gabriel.santos, esther,sandra\}@ic.unicamp.br}
}

\date{}

\begin{document}
\maketitle
\begin{abstract}

In-context learning (ICL) enables Large Language Models (LLMs) to learn tasks from demonstration examples without parameter updates. Although it has been extensively studied in LLMs, its effectiveness in Vision-Language Models (VLMs) remains underexplored. In this work,~we present a systematic study of ICL in VLMs, evaluating seven models spanning four architectures on three image captioning benchmarks. We analyze how prompt design, architectural choices, and training strategies influence multimodal ICL. To our knowledge,  we are the first to analyze how attention patterns in VLMs vary with an increasing number of in-context demonstrations.
Our results reveal that training on image–text interleaved data enhances ICL performance but does not imply effective integration of visual and textual information from demonstration examples. In contrast, instruction tuning~impro\-ves instruction-following but can reduce reliance on in-context demonstrations, suggesting a trade-off between instruction alignment and in-context adaptation. Attention analyses further show that current VLMs primarily focus on textual cues and fail to leverage visual information, suggesting a limited capacity for multimodal integration. These findings highlight key limitations in the ICL abilities of current VLMs and provide insights for enhancing their ability to learn from multimodal in-context examples.

\end{abstract}

\section{Introduction}
\label{sec:introduction}

Large Language Models (LLMs) have demonstrated notable performance in a wide range of Natural Language Processing (NLP) tasks, showcasing their potential across various domains. As the scale of LLM increases, in-context learning (ICL) emerges as a new ability that enables models learning new tasks from a few demonstration examples~\cite{brown2020language, wei2022emergent}. In this paradigm, the text generation process is conditioned on a set of input-output, i.e.,  demonstrations, which enhance the prompt with contextual information. Since ICL does not require parameter updates~\cite{dong2024survey}, it has been used as a cost-effective alternative to traditional fine-tuning for many NLP applications.

Although the ICL ability of LLMs has been studied from multiple perspectives~\cite{dong2024survey}, comparatively little attention has been given to understanding this capacity in Vision-Language Models (VLMs)~\cite{baldassini2024makes, qin2024what, yang2024exploring, chen2024understanding}. Exploring ICL in VLMs is particularly important because strategies that are effective for LLMs are not necessarily transferable to multimodal settings, as demonstrated by~\citet{li2024configure}. Previous works have mostly focused on investigating demonstration selection and ordering strategies as well as the contribution of each modality to ICL, with an emphasis on tasks such as Visual Question Answering (VQA) and image classification. However, these studies typically evaluate a limited set of models trained on interleaved image-text data (i.e., datasets composed of instances consisting of multiple images and texts interleaved), leaving open questions about the generalization of their findings to other tasks and to VLMs trained on image-text pair datasets, where each instance comprises only an image and an associated text. In addition, it remains unclear how VLMs use their contextual information when performing downstream tasks.

To address this gap, we present a comprehensive study of ICL in VLMs, with a focus on the task of image captioning. We evaluate seven models that cover four distinct architectures across three image captioning benchmarks. Unlike prior work, our analysis includes both models trained on interleaved image-text data (OpenFlamingo~\cite{awadalla2023openflamingo}, Idefics2~\cite{laurencon2024what_matters}, and LLaVA-Next-Interleave~\cite{li2024llava_next}) and models trained on image-text pair datasets (InstructBLIP~\cite{dai2024instructblip} and LLaVA v1.5~\cite{liu2023improved}). Through a series of controlled experiments, we systematically analyze how different model architectures and training strategies impact ICL performance. Particularly, we conducted experiments varying the instructions, blacking out and removing demonstration images, and we studied their effect on the ICL capacity of different VLMs. We hypothesize that a VLM with robust multimodal ICL capacity can efficiently leverage the textual and visual information from demonstration examples to generate the answer. Then, its performance should be minimally impacted by the changes in instruction. Still, it should be significantly hampered when corrupting demonstrations (in this case, removing or blacking out images). To our knowledge, we are the first to investigate ICL in VLMs through the lens of attention patterns, offering new insights into how VLMs attend to context and revealing limitations in their current ICL capabilities.

Our main findings are as follows:
\begin{itemize} 
    \item Training data structure significantly impacts ICL capacity; in particular, training on image-text interleaved datasets improves models' ICL ability. However, this benefit does not imply effective integration and use of visual and textual information from demonstration examples.\vspace{-0.2cm}
\item Through an analysis of attention maps, we find that the evaluated models do not fully exploit in-context visual information; their ICL behavior is primarily driven by textual context, suggesting a limited integration of multimodal cues.\vspace{-0.2cm}
\item While instruction tuning improves instruction-following ability, allowing models to comprehend detailed instructions, it can impair ICL by diminishing the model's reliance on in-context demonstration.
\end{itemize}
These findings highlight crucial limitations in current VLMs that should be addressed to enhance their multimodal ICL ability.

\section{Related Work}
\label{sec:related_work}

\paragraph{VLMs.}
VLMs excel in vision-language tasks due to pre-trained visual encoders and LLMs~\cite{yin2023survey, zhang2024survey}. They comprise three key components: a visual encoder for image features, an LLM for text generation, and a modality projector to align visual and textual data, bridging the modality gap.

Various approaches have been explored for the modality projector, including linear layers and multi-layer perceptrons (MLPs)~\cite{koh2023fromage, liu2023improved, shukor2023epalm, su2023pandagpt, lin2024vila, liu2024visual}, which, despite the low training costs, can lead to long sequences of tokens, thereby increasing the inference costs. Pooling strategies help mitigate this issue~\cite{cha2024honeybee, sun2024generative, hu2024mplug}. Advanced methods like Q-Former~\cite{blip2_li_2023} improve alignment between frozen visual encoders and LLMs~\cite{zhu2024minigpt, dai2024instructblip, geigle2024mblip}. Another alternative is to use interleaved cross-attention layers~\cite{alayrac2022flamingo, laurencon2023obelics, xue2024blip3}, in which the LLM directly attends to visual features. However, this approach substantially increases the number of trainable parameters, as noted by~\citet{laurencon2024what_matters}.

Training these models typically involves pretraining the modality projector on large-scale image-text datasets while keeping the visual encoder and LLM frozen for feature alignment. Subsequently, the LLM can be fine-tuned alongside the modality projector on instruction-following datasets to improve zero-shot generalization. Most works~\cite{dai2024instructblip, liu2024visual, liu2023improved, zhu2024minigpt, hu2024mplug} train on a mixture of image captioning~\cite{MSCoco2014, li2022blip, sharma2018conceptual}, VQA~\cite{goyal2017making, schwenk2022okvqa, marino2019ok}, and instruction-following~\cite{liu2024visual} datasets. Some models, such as Flamingo~\cite{alayrac2022flamingo}, Idefics~\cite{laurencon2023obelics, laurencon2024what_matters, laurencon2024buildingbetter}, VILA~\cite{lin2024vila}, MMICL~\cite{zhao2024mmicl}, MM1~\cite{mckinzie2025mm1}, and xGen-MM (BLIP-3)~\cite{xue2024blip3}, are trained on interleaved image-text datasets~\cite{laurencon2023obelics, zhu2024multimodal} to further enhance multimodal reasoning capabilities.

\paragraph{ICL in VLMs.}~Although ICL has been widely studied in LLMs, it remains relatively underexplored in VLMs. Recent works have investigated the factors that influence ICL in VLMs, including modality importance, recency bias, demonstration retrieval, and ordering strategies. However, these studies are generally limited to a small set of models trained on interleaved image-text datasets, with a focus primarily on VQA and image classification~tasks.

\citet{qin2024what} studied ICL in VLMs trained with interleaved data under different scenarios. They showed that the internal order of the modalities within each demonstration has a greater impact on performance than the arrangement of demonstrations themselves. Also, unlike ICL in LLMs, where increasing the number of demonstrations typically improves performance, they did not observe significant performance gains when providing more demonstrations.

\citet{yang2024exploring} investigated ICL for image captioning, analyzing different strategies for demonstration retrieval and caption assignment. Their findings suggest that when demonstration images are similar to the query image, VLMs may leverage in-context captions as shortcuts to generate a new one rather than learning the captioning task. They conducted their experiments, however, within a restricted scope, using only \text{MS COCO}~\cite{MSCoco2014} and experimenting with only Idefics and OpenFlamingo models, which limited the generalizability of their conclusions.

More related to our work, \citet{chen2024understanding} and~\citet{baldassini2024makes} investigated ICL in two Flamingo-based VLMs: Idefics and OpenFlamingo. They showed that textual information plays a more important role than visual information in the demonstrations. Removing images results in only a minor performance decrease, whereas corrupting textual descriptions leads to a significant performance decline, indicating that these VLMs heavily rely on textual cues even when processing multimodal demonstrations. Moreover, \citet{baldassini2024makes} found that these models exhibit recency bias, tending to replicate outputs of the most recent demonstrations, even when earlier demonstrations are more semantically relevant.

In this work, we focus on the task of image captioning and present a systematic analysis of ICL in seven VLMs across four distinct architectures and three benchmark datasets. Unlike previous studies, we extend our investigation to include InstructBLIP~\cite{dai2024instructblip} and LLaVA v1.5~\cite{liu2023improved}, originally designed for single image-text pairs. To our knowledge, this is the first comprehensive evaluation of ICL in VLMs that have not been trained on interleaved image-text datasets. Additionally, we are the first to investigate attention patterns across the layers of text decoder blocks in different VLM architectures as the number of demonstrations varies, providing new insights into the limits of their ICL capacity.

\section{Methodology}
\label{sec:methodology}

\subsection{Experimental Setup}
\label{sub:setup}

\paragraph{Models.}
We analyze four distinct families of VLMs: InstructBLIP~\cite{dai2024instructblip}, LLaVA ~\cite{liu2023improved, li2024llava_next}, OpenFlamingo~\cite{awadalla2023openflamingo}, and Idefics2~\cite{laurencon2024what_matters}. These families were selected to systematically explore how various design choices, such as bridging the modality gap and different training strategies, affect the ICL capabilities of VLMs. 

We use model checkpoints with parameter sizes ranging from 4B to 9B for a fair comparison across similar scenarios. Specifically, for InstructBLIP, we evaluate two checkpoints with different LLMs: InstructBLIP FlanT5-XL and InstructBLIP Vicuna~7B. For the other families, we assess LLaVA~v1.5~7B, LLaVA-NeXT-Interleave, OpenFlamingo~9B, and two checkpoints of Idefics2, before and after the instruction-tuning phase,  namely, Idefics2 (Base) and Idefics2 (IT)\footnote{
\url{https://huggingface.co/Salesforce/instructblip-flan-t5-xl},
\url{https://huggingface.co/Salesforce/instructblip-vicuna-7b},
\url{https://huggingface.co/llava-hf/llava-1.5-7b-hf},
\url{https://huggingface.co/llava-hf/llava-interleave-qwen-7b-hf},
\url{https://huggingface.co/openflamingo/OpenFlamingo-9B-vitl-mpt7b},
\url{https://huggingface.co/HuggingFaceM4/idefics2-8b-base}, and
\url{https://huggingface.co/HuggingFaceM4/idefics2-8b}.
}. 

\paragraph{Datasets \& Metrics.}
We evaluate the models using three image captioning benchmarks: \text{MS COCO}~\cite{MSCoco2014}, Flickr30K~\cite{young2014flickr30k} and NoCaps~\cite{agrawal2019nocaps} datasets.
We conduct our evaluation on the respective validation sets, utilizing the MS COCO training set as the knowledge base from which we retrieve similar examples to construct the context. Each demonstration example comprises an image-caption pair, with the caption being randomly sampled from the multiple human annotations available for MS COCO. We employ CIDEr-D~\cite{vedantam2015cider} as the evaluation metric.

\subsection{Evaluation Protocol}
\label{sub:evaluation_protocol}

\paragraph{Demonstrations Retrieval.} 
Inspired by~\citet{yang2023re}, we retrieve demonstration examples employing a $k$-Nearest Neighbor approach based on the similarity distance in the visual feature space. We construct a knowledge base $\mathcal{D} = \{(i_1, t_1), \ldots , (i_n, t_n)\}$, consisting of images~$i$ paired with their corresponding texts $t$ different from those in the evaluation sets. In our experiments, we use the MS COCO training set as our knowledge base.

\begin{figure*}[t]
    \centering
    \includegraphics[width=0.8\textwidth]{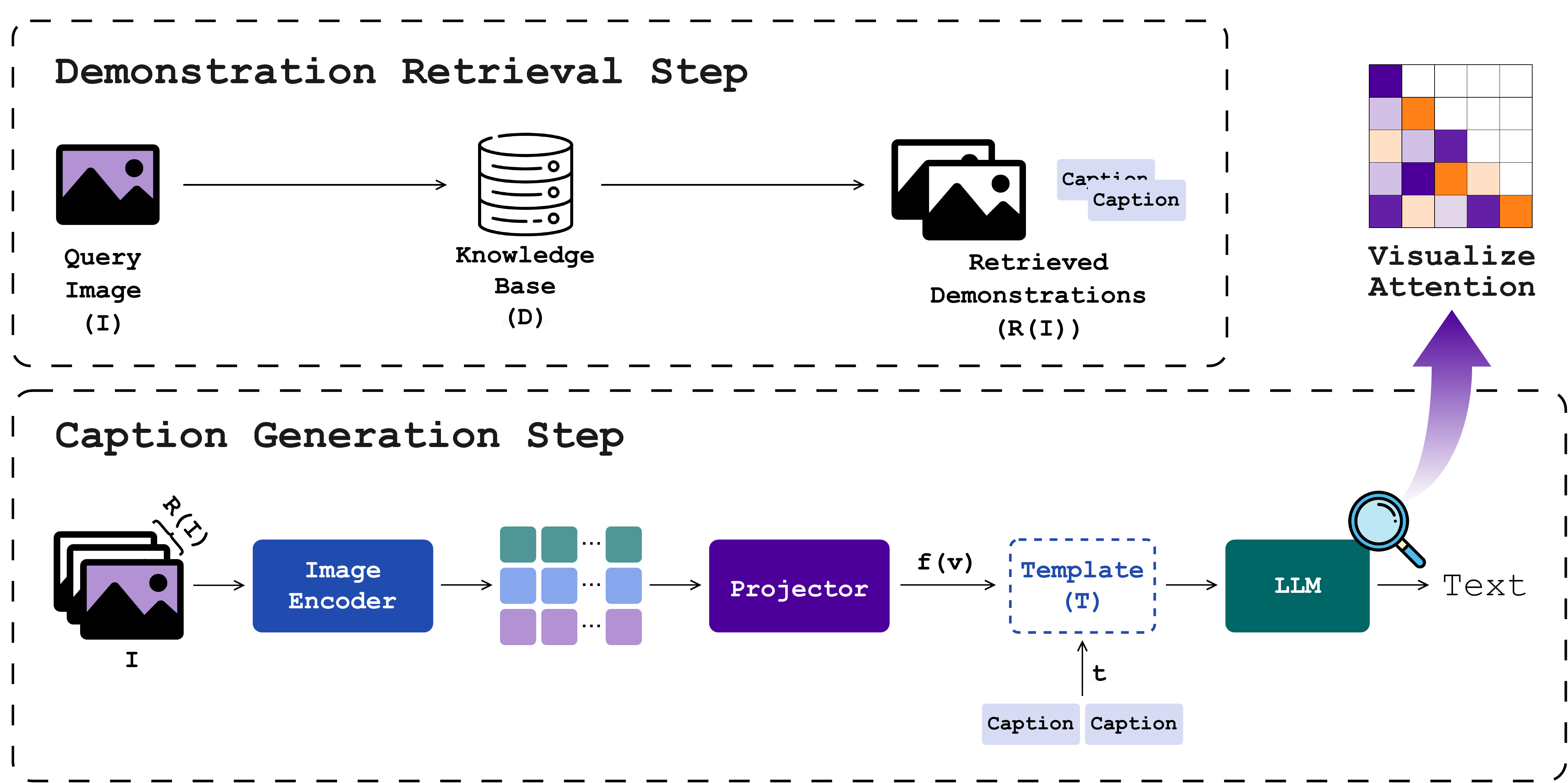}
    \caption{
    Overview of our evaluation pipeline for assessing the ICL capability of VLMs. We illustrate the demonstration retrieval and caption generation steps.}
    \label{fig:multimodal_icl_pipeline}

\end{figure*}

For each query image $I$, we extract its features $f(I)$ and we retrieve the top-$k$ most similar image-text pairs based on the cosine similarity between visual features, as illustrated in the ``Demonstration Retrieval Step'' in Figure~\ref{fig:multimodal_icl_pipeline}. Formally, the retrieved set $\mathcal{R}(I)$ of image-text pairs is defined as $\mathcal{R}(I) = \{(i, t) \, |\,  \text{top-}k_{(i, t) \in \mathcal{D}} \ sim(f_I, f_i)\}$\footnote{For simplicity, we denote $f(i)$ as $f_i$ and  $f(I)$ as $f_I$.}, where $sim(\cdot)$ denotes the cosine similarity. We use a ViT~\cite{dosovitskiy2021an}\footnote{\url{https://huggingface.co/google/vit-large-patch16-224-in21k}} to encode the \text{images}. 

\paragraph{ICL.} 
During the caption generation step (Figure~\ref{fig:multimodal_icl_pipeline}), we first extract visual features from the query image and from the images in the retrieved demonstration set $\mathcal{R}(I)$. These features are projected into the LLM's token embedding space, producing visual tokens denoted as $f(v)$. We then construct a multimodal prompt by inserting the visual tokens and their corresponding captions into a predefined template $\mathcal{T}$. This prompt is passed to the LLM to generate the caption.

Although this pipeline is implemented in a relatively straightforward manner for LLaVA, Idefics2, and OpenFlamingo, adapting it to InstructBLIP presents additional challenges. InstructBLIP employs a Q-Former module to extract instruction-aware visual features. To extend it to handle multiple demonstrations, we process each demonstration image independently using the Q-Former, paired with a fixed instruction: \texttt{``a short image description''}. The resulting visual query tokens are then embedded in the template $\mathcal{T}$ alongside their corresponding captions.

\paragraph{Templates.}\label{sub:methodology_templates} To evaluate the models' ability to adapt at inference time, we experiment with two different templates. First, for each model, we construct a straightforward template based on its original training instructions, into which we directly embed the demonstration examples $\mathcal{R}(I)$, i.e., image-caption pairs.

\begin{figure}[t]
    \centering
    \includegraphics[width=1\linewidth]{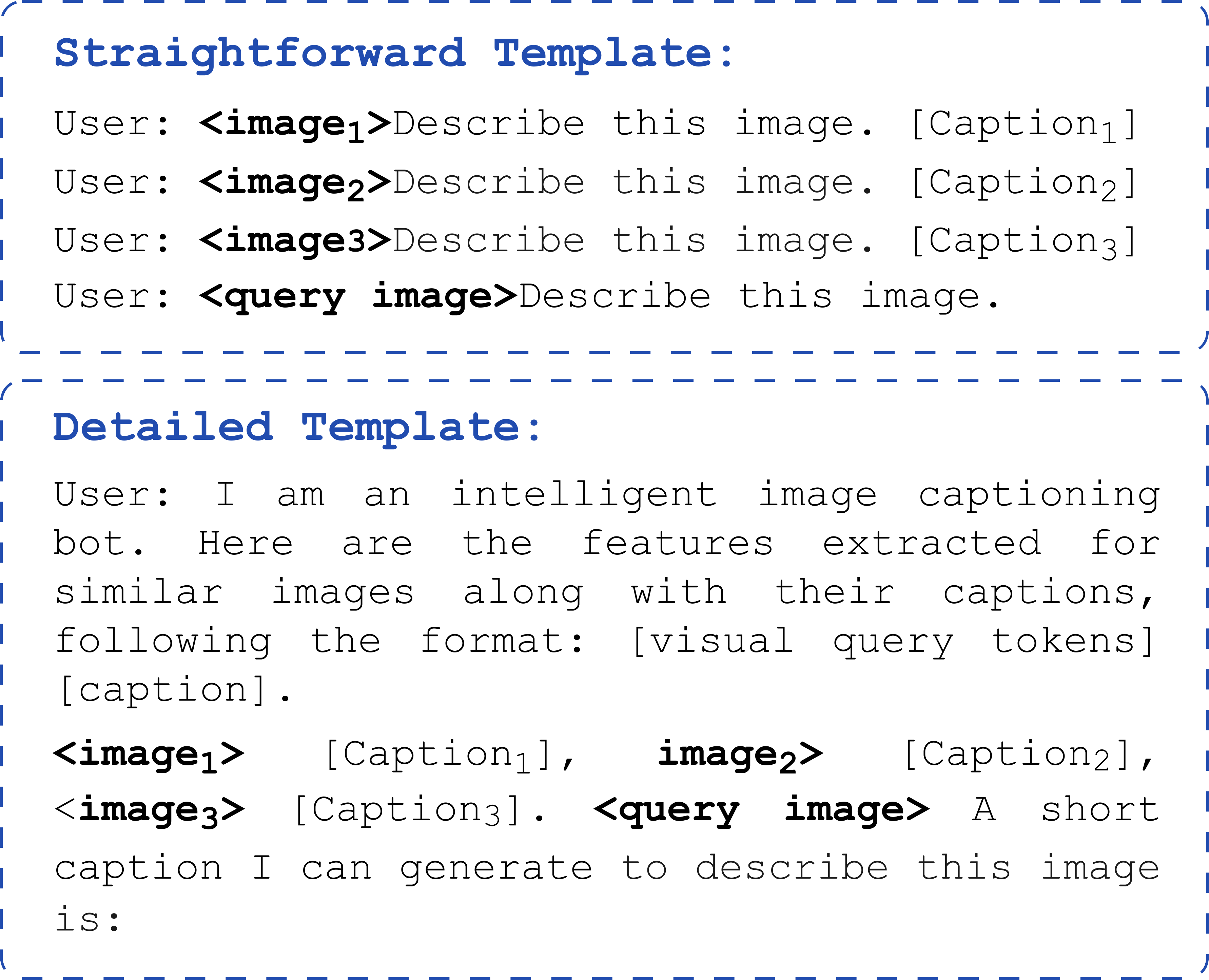}
    \caption{Investigated templates.}
    \label{fig:prompts}
\end{figure}

Additionally, building upon the Socratic Models, we further explore detailed prompts based on Socratic templates~\cite{zeng2023socratic, ramos2023lmcap} that specify the task and the format for presenting demonstration examples (Figure~\ref{fig:prompts}). Following~\citet{baldassini2024makes}, in our experiments, we sort the demonstrations in increasing order of similarity to the query image, as the models tend to assign greater importance to the last demonstrations. 

\section{Results and Discussions}
\label{sec:results}

\paragraph{Effect of Prompt Structure on ICL.}
\label{sub:results_icl_prompt}
\begin{figure*}[!htb]
    \centering
    \includegraphics[width=\textwidth]
    {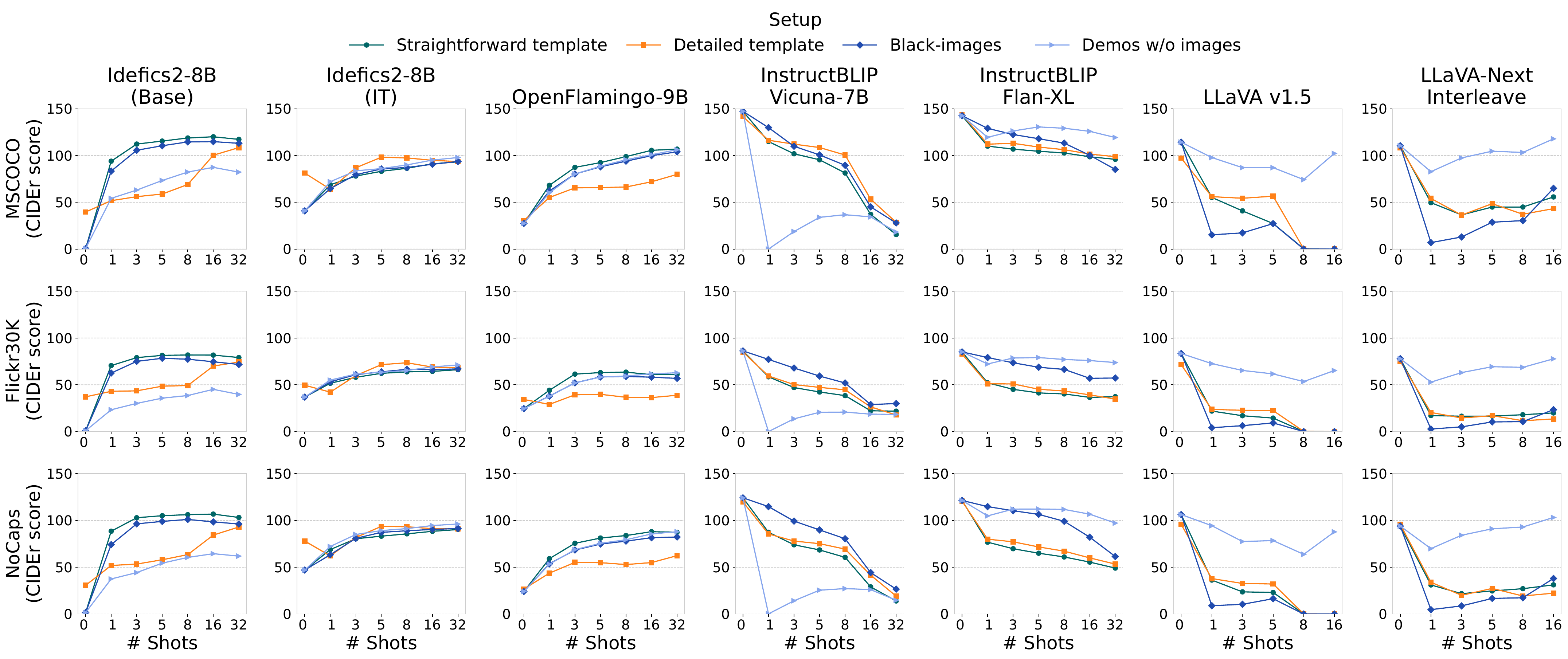}
     \caption{\textbf{Evaluating ICL capacity of VLMs across different scenarios}. We evaluate the models using both straightforward and detailed templates. Additionally, we explore scenarios where demonstration captions are provided. However, the demonstration images are blacked out, as well as cases where only the captions are available as context. ``Idefics2 8B (IT)'' refers to the instruction-tuned checkpoint of the Idefics2 architecture.}
    \label{fig:icl}
\end{figure*}

To investigate the influence of prompt structure on ICL, we evaluate models on the image captioning task using prompts designed with two levels of detail. The first is built using the straightforward template, where the demonstration image-caption pairs are simply concatenated with an instruction. The second employs the detailed template, clearly specifying the format of the examples and including the phrase ``\texttt{I am an intelligent image captioning bot}'' (Figure~\ref{fig:prompts}). Figure~\ref{fig:icl} shows the results.

Overall, we observed that Idefics2 (Base) and OpenFlamingo perform better when using the straightforward template compared to the detailed one. However, as we increase the number of demonstrations (shots), Idefics2 (Base) begins to perform similarly across both templates. This trend does not hold for OpenFlamingo, which consistently achieves higher scores with the straightforward template regardless of the number of shots. Interestingly, after the instruction-tuning step, Idefics2 exhibits similar performance for both templates. Also, a closer look reveals that Idefics2~(IT) demonstrates significant gains in the zero-shot scenario compared to Idefics2 (Base). However, it converges to a point below that of its base checkpoint, indicating a deterioration of ICL. 

InstructBLIP models and LLaVA models\footnote{Due to computational constraints, we evaluate LLaVA models up to 16 demonstrations.}, both instruction-tuned models, perform similarly with straightforward and detailed templates. 
 These results align with previous works~\cite{liu2024visual, wu2024language}, indicating that instruction tuning enhances a model's ability to follow instructions, as evidenced by similar performance across different prompts after fine-tuning. On the other hand, our results suggest that it can also significantly hamper the model's ICL ability, as seen in Idefics2 models, where CIDEr-D scores drop by 20~points after instruction tuning.

\paragraph{Interleaved vs.~Paired Image-Text Training.}
\label{sub:results_training_strategies}
Additionally, we investigate how the ICL capacity of VLMs trained on image-text interleaved datasets (Idefics2, OpenFlamingo, and LLaVA-Next-Interleave) differs from those trained on datasets composed of image-text pairs (InstructBLIP and LLaVA v1.5). As shown in Figure~\ref{fig:icl}, these two training paradigms yield opposite trends as the number of demonstrations increases. Models trained on image-text interleaved datasets perform consistently better with more demonstrations, indicating strong ICL capabilities. In contrast, the performance of models trained on image-text paired datasets significantly declines as the number of shots increases, showing they have a limited ICL capacity. Notably, LLaVA v1.5's performance drops to zero when given eight or more demonstrations, whereas LLaVA-Next-Interleave shows stable performance across the number of demonstrations.

This finding suggests that the training data structure plays a critical role in shaping a model's ICL capacity, with image-text interleaved datasets contributing to enhancing such capacity. Furthermore, we observe that the two variants of InstructBLIP behave differently. While InstructBLIP Flan-XL appears to plateau at eight shots, InstructBLIP \text{Vicuna-7B} continues to decline. We hypothesize that this difference arises from the presence of few-shot templates in the Flan-T5 training set, which enhances the LLM's ICL capabilities in the textual domain. Then, part of the ICL ability learned by the VLM's text decoder can also be leveraged in the multimodal setup.

\paragraph{Do VLMs ``See'' In-Context Images?}
\label{sub:results_modalities}

To investigate the contribution of the visual modality to model performance, we evaluate two ablation scenarios. First, given a query image, we retrieve similar demonstrations but replace the retrieved images with black ones while preserving their captions.  These modified demonstrations are then inserted into the straightforward template (Figure~\ref{fig:prompts}). In the second scenario, rather than blacking out the demonstration images, we simply remove them from the context, including only their associated captions into the straightforward template. The results of these experiments are shown in Figure~\ref{fig:icl}.

Comparing the performance of models using demonstrations with original images (straightforward template) against those with blacked-out images (black-images), we find that most models perform similarly across different shot counts. Particularly, both InstructBLIP variants exhibit improved performance on Flickr30K and NoCaps when images are blacked out, although still with a downward trend with respect to shot. In contrast, when we remove demonstration images from the context (demos w/o images), models behave differently. Idefics2 (Base) and InstructBLIP Vicuna-7B exhibit a sharp performance drop, especially at low shot counts. For Idefics2 (Base), this performance degradation is likely because when we pass only similar captions as context, we disrupt the image-text interleaved structure on which it was originally trained, resulting in a prompt out of the training distribution and confusing the model.  This issue seems to be mitigated in Idefics2 (IT), indicating that instruction tuning also enhances robustness to such structural changes. OpenFlamingo, in turn, does not exhibit a significant difference in performance when using or not using demonstration images. Conversely, LLaVA models and InstructBLIP Flan-XL perform better when only captions are included, suggesting that the ICL capacity of these models relies mostly on text while visual content may distract their LLMs during text decoding. Overall, these findings suggest that the evaluated models do not ``see'' images in the context; instead, their ICL ability is predominantly based on the textual modality.

\paragraph{Analysis of Attention Patterns.}
\label{sub:results_attention}

\begin{figure*}[!htb]
    \centering
    \includegraphics[width=1\textwidth]{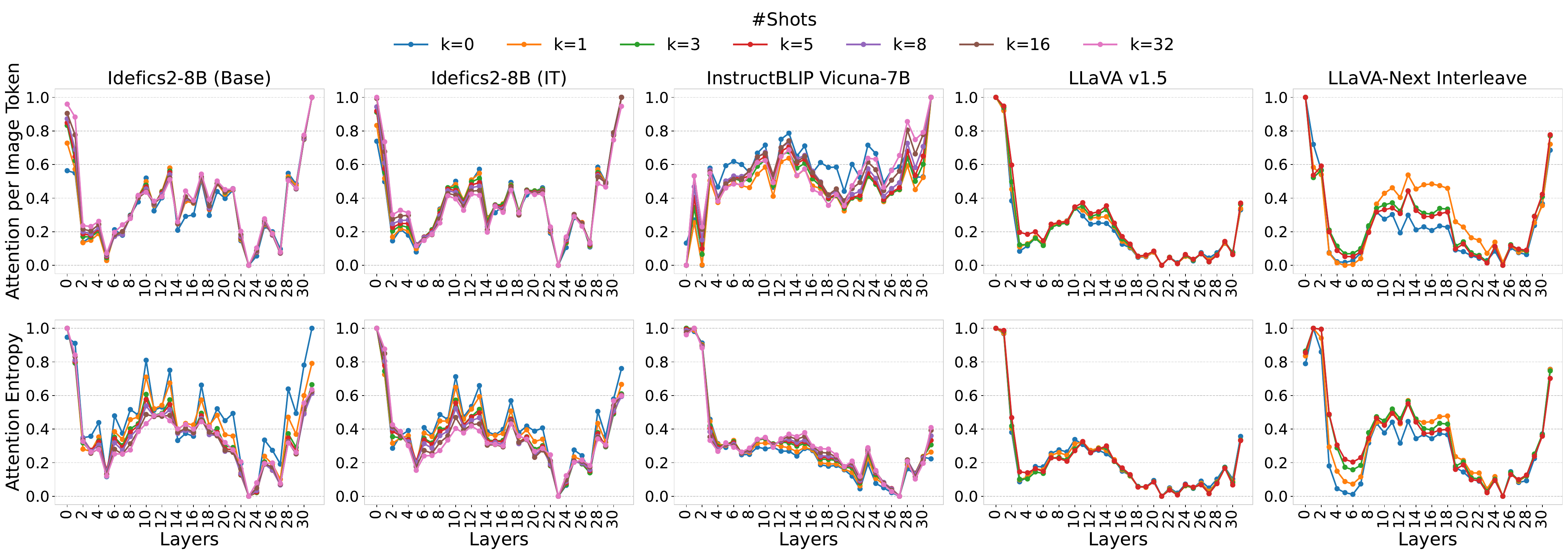}
    \caption{\textbf{Layer-wise attention analysis}.
The upper row presents the variation of mean attention weight assigned to a visual token across the models' LLM layers. The lower row shows the attention entropy across all tokens at each LLM layer, reflecting the diffuseness of attention distribution. For comparability, the charts plot min-max normalized values.}
    \label{fig:attention_across_layers}
\end{figure*}

\begin{figure*}[!htb]
    \centering
    \includegraphics[width=1\textwidth]{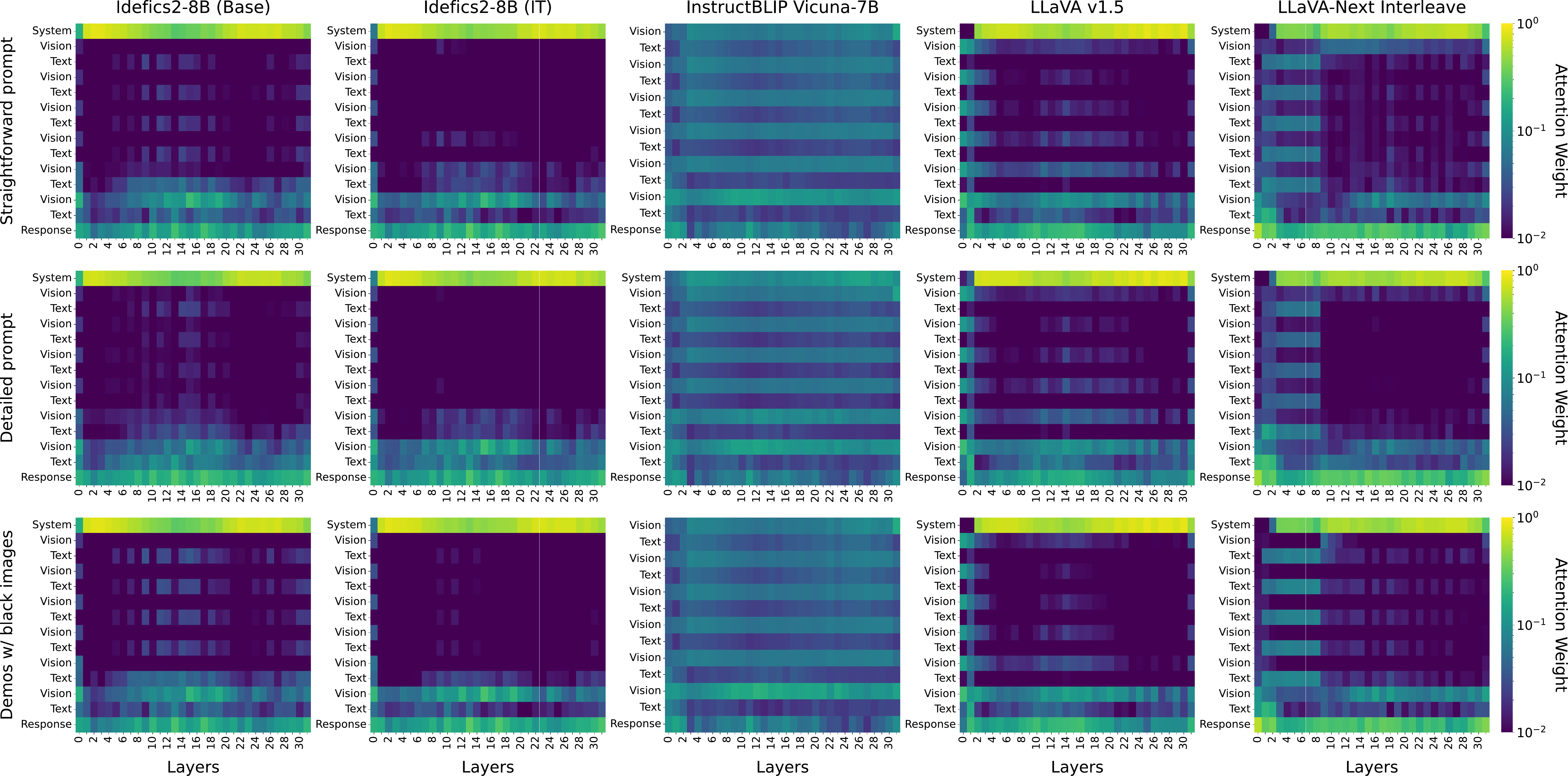}
    \caption{\textbf{Attention maps with scores aggregated by token type} (log-scale). Maps are plotted for InstructBLIP Vicuna-7B and Idefics2 models, comparing the 5-shot setting across prompts built on the straightforward template (first row), detailed template (second row), and demonstrations with blacked-out images (third row). Columns correspond to the respective models.}
    \label{fig:attention_maps}
\end{figure*}

To further analyze what models ``see''  and how different training strategies impact ICL capacity, we select Idefics2 models, InstructBLIP Vincuna-7B, LLaVA v1.5, and LLaVA-Next-Interleave for a close analysis. The choice of these models took into account the similarity in terms of architecture of these models, as both of them use decoder-only LLMs and pass the visual information as input tokens to the LLM. We investigate how the attention weights assigned to visual tokens and the entropy of attention across tokens vary throughout the LLMs layers and whether the patterns are consistent as the number of in-context demonstrations increases. In this experiment, we use the straightforward template.

Figure~\ref{fig:attention_across_layers} shows that overall models exhibit similar trends for the attention entropy. It is maximum in the lower layers, then there is a sharp fall followed by an inverted U-shaped curve in the middle layers, and finally it rises again. Moreover, considering normalized values, entropy trends remain relatively stable across different shot counts. However, Idefics2 (Base) presents higher entropy in the final layers under the zero- and one-shot condition, consistent with its weaker performance in those settings.  The results show that, for all evaluated models, attention is more diffuse among tokens in early and also in late layers for Idefics2 models.

In terms of attention per image token, the variation of normalized attention scores across layers shows a similar pattern for all shots, consistently observed across all evaluated models. Moreover, models assign maximum attention to the visual tokens in the early layers with a sharp fall followed by an inverted U-shaped curve in the middle layers, and finally it turns to an increase in the later layers. Conversely, InstructBLIP Vicuna-7B assigns minimal attention to visual tokens in early layers, which increases in deeper layers. This difference can be explained by the fact that InstructBLIP Vicuna-7B was the only model that kept the text decoder frozen during the whole training. Then, the LLM possibly treats the visual tokens in the same way as textual tokens. On the other hand, when the LLM is unfrozen during VLM's training, it may learn to assign higher attention to visual tokens in early layers in order to extract relevant visual information. This finding is consistent with the previous conclusion~\cite{zhang2025llavamini}, and we demonstrate that it also generalizes to Idefics2 models, in addition to LLaVA, and holds across different shot numbers.

Next, we further analyze how the models attend to individual images and text segments in the context. We plot the total attention weight assigned to each subsequence, i.e., textual or visual tokens, and previously generated tokens (response), across all layers. We conduct this analysis in the 5-shot setting, using straightforward and detailed templates, as well as demonstrations with blacked-out images.

As shown in Figure~\ref{fig:attention_maps}, InstructBLIP Vicuna-7B's behavior contrasts with Idefics2 and LLaVA models. InstructBLIP Vicuna-7B distributes attention more uniformly across the tokens in the early layers, also evidenced by the high entropy, and concentrates attention on visual tokens in the middle and final layers. In contrast, Idefics2 and LLaVA models assign the highest attention to the first textual segment, system prefix\footnote{All models except InstructBLIP Vicuna-7B use a system prompt (e.g., \texttt{``USER:''}) before user instruction.}, across all layers. Moreover, these models tend to concentrate attention near the end of the token sequence, particularly on the query image, task instruction, and previously generated tokens. This observation aligns with the conclusion of~\citet{liu2024lost} that LLMs give more importance to information at the beginning and end of context, and instruction fine-tuned models tend to assign a high attention score to the system prefix. It is also consistent with the improved performance reported when demonstrations are sorted by increasing similarity to the query image~\cite{baldassini2024makes}.

Moreover, LLaVA v1.5 assigns higher attention to visual tokens in early layers and less significant attention in the middle layers, while ignoring the textual information of demonstrations. Conversely, LLaVA-Next-Interleave assigns insignificant attention to demonstration images and higher attention to textual tokens of demonstrations in early and middle layers, focusing on the query image and system prefix in the late layers. Idefics2 (Base), in turn, assigns attention to the query and the last few images in early layers, and it distributes attention among demonstration captions in the middle and final layers. After instruction tuning, Idefics2 (IT) seems to ignore the information in the middle of the context after the first layer. This is consistent with the weaker ICL ability of Idefics2 (IT) compared to Idefics2 (Base), suggesting that instruction tuning may reduce the use of demonstration content. We observe no substantial differences in attention distribution between the straightforward and detailed templates, nor between original and blacked-out demonstrations.

These findings further support our hypothesis that the evaluated VLMs have limited capacity to leverage multimodal in-context information. For Idefics2 and LLaVA-Next-Interleave models, ICL appears to rely predominantly on textual information. Moreover, instruction tuning in Idefics2 may reduce reliance on demonstration captions, potentially impacting ICL performance. In contrast, InstructBLIP Vicuna-7B concentrates attention on images while ignoring the captions. These insights underscore the importance of achieving balanced attention across modalities to fully exploit ICL potential in multimodal settings.

\section{Conclusions}
\label{sec:conclusion}

In this paper, we conduct a comprehensive study of ICL in VLMs, evaluating seven models spanning four distinct architectures on several image captioning benchmarks. We investigate how prompt design, model architecture, and training data structure influence ICL performance. In contrast to prior work, we go beyond models trained solely on interleaved image-text data; we also analyze VLMs trained on datasets composed of image-text pairs. We find that instruction tuning can enhance instruction-following ability, allowing models to comprehend detailed instructions, but it can hamper ICL capacity by diminishing the model's reliance on in-context demonstration. In contrast, training on interleaved image-text datasets enhances ICL ability. However, the benefits do not necessarily extend to multimodal settings. Our attention map analysis reveals that these models do not fully leverage visual inputs; their ICL behavior is largely driven by textual information, indicating limited capacity for integrating multimodal information.

Our findings uncover critical limitations in current VLMs. Future research should investigate whether our findings generalize to larger-scale models and explore to what extent VLMs inherit and utilize the ICL capabilities of their underlying LLMs. Designing more effective modality projectors may be the key to better transferring these abilities. Finally, investigating training strategies that combine instruction tuning with interleaved image-text supervision to support both instruction following and contextual learning is a promising direction.

\section*{Limitations}
\label{sec:limitations}

Although our analysis focuses on VLMs with up to 9B parameters due to computational constraints, studying larger models would be important to determine whether our conclusions hold at a greater scale. Furthermore, to better understand the role of instruction-tuning and training of interleaved image-text datasets, it would be interesting to evaluate more models before and after instruction-tuning. Finally, our analysis is limited to VLMs trained in English-language texts. However, evaluating the ICL capacity of multilingual models is essential. It would be necessary to study whether ICL can improve VLMs' performance on low-resource languages.

\section*{Ethics Statement}
\label{sec:ethics_statement}

This study systematically analyzes the ICL capabilities of publicly available VLMs. Our analysis is based solely on publicly available image captioning datasets, and we fully comply with the terms of use and licensing agreements associated with each model and dataset. We do not conduct any fine-tuning or modifications in the models that could introduce unintended risks. However, we recognize that our work reflects the existing limitations and potential risks of the evaluated models, including, but not limited to, gender, racial, and cultural biases, as well as the potential for generating misinformation or disinformation.


\bibliography{custom}
\bibliographystyle{tacl_natbib}

\newpage
\appendix
\section{Appendix}
\label{sec:appendix}

\setcounter{table}{0}
\renewcommand{\thetable}{A\arabic{table}}

\setcounter{figure}{0}
\renewcommand{\thefigure}{A\arabic{figure}}


\subsection{Details on Experimental Setup}
To facilitate the reproducibility of our work, we report in Table~\ref{tab:models_info} the models we analyzed, along with details on their number of parameters and training dataset size, as well as the energy consumption and carbon emissions from our experiments estimated with \texttt{codecarbon}~\cite{codecarbon}. Table~\ref{tab:datasets_info} shows the datasets used in our experiments, including statistics on their size. Additionally, we outline the main text decoding hyperparameters used in our experiments in Table~\ref{tab:hyperparams_image_captioning}. Note that we use the hyperparameters reported for each model for the image captioning task. However, LLaVA~v1.5 does not report hyperparameters for this task, then we use the ones from InstructBLIP, which led to the best results in our preliminary experiments. We conducted our experiments in a heterogeneous infrastructure; however, the majority were performed on a single NVIDIA A100 80GB GPU. 

\begin{table}[!htb]
\setlength{\tabcolsep}{0.9mm}
\caption{VLMs investigated in this work. For each model, we report the number of parameters, the size of the training dataset, and the estimated energy consumption and carbon emissions from our experiments.}
\label{tab:models_info}
\resizebox{\columnwidth}{!}{%
\begin{tabular}{lcccc}
\midrule
Model & \begin{tabular}[c]{@{}c@{}} \#Params \\ (B) \end{tabular} & \begin{tabular}[c]{@{}c@{}} Training \\ Set Size (M) \end{tabular} & \begin{tabular}[c]{@{}c@{}} Energy \\ (kWh) \end{tabular} & \begin{tabular}[c]{@{}c@{}}Emission \\ ($\text{CO}_{2}$eq in kg)\end{tabular} \\
\midrule
Llava v1.5-7B & 7.1 & 0.15 & 504.4 & 1.2 \\
InstructBLIP Vicuna-7B & 7.9 & 15.1 & 143.1 & 14.1 \\
InstructBLIP Flan-XL & 4.0 & 15.1 & 43.4 & 0.1 \\
Idefics2-8B & 8.4 & 351.2 & 52.8 & 5.2 \\
OpenFlamingo-9B & 8.1 & 2,101.0 & 1,285.0 & 126.4 \\
\midrule
\end{tabular}%
}
\end{table}

\begin{table}[!htb]
\setlength{\tabcolsep}{4mm}
\caption{Datasets used in our experiments. For each dataset, we report the number of samples in each split and the specific task it is used for.  Note that we do not use Flickr or NoCaps training sets, as we rely on the MS COCO training set as the knowledge base for these datasets. ``Val.'' stands for the validation dataset.}
\label{tab:datasets_info}
\scriptsize
\centering
\begin{tabular}{lcc}
\midrule
\multicolumn{1}{l}{Dataset} & \multicolumn{1}{c}{Size} \\ \midrule
MS COCO & Train: 118.2K/ Val: 5.0K \\
Flickr30K & Val: 1.0K  \\
NoCaps & Val: 4.5K  \\
\midrule
\end{tabular}%
\end{table}

\begin{table}[!htb]
\setlength{\tabcolsep}{1.7mm}
\caption{Text decoding hyperparameters.}
\label{tab:hyperparams_image_captioning}
\resizebox{\linewidth}{!}{%
\centering
\begin{tabular}{lccc}
\midrule
Hyperparameters & InstructBLIP / LLaVA & Idefics2 & OpenFlamingo \\
\midrule
\# Beams & 5 & -- & 3 \\
Max. New Tokens & 30 & 20 & 20 \\
Min. Length & 10 & -- & -- \\
Repetition Penalty & 1.5 & -- & -- \\
Length Penalty & 1.0 & -- & -- \\
\midrule
\end{tabular}%
}
\end{table}

\subsection{Formalization}

In Section~\ref{sub:results_attention}, we analyze the attention patterns in different VLMs. Here, we formalize how we aggregate the attention scores in our experiments.

\subsubsection{Attention per Token Type}
Formally, given $a^{(h, d)}_{i, j}$ the attention score from $i$-th to $j$-th token at $h$-th attention head for the $d$-th sample, a sequence of textual and visual tokens $T=(t_1, v_1, \ldots , t_k, v_k)$ and a sequence of response tokens $R$, we define the attention score per token type $T_k$ as follows:
\begin{equation}
\label{eq:avg_attention}
\bar a_{i, j} = \frac{1}{|\mathcal{D}|} \frac{1}{H} \sum_{d} \sum_{h} a^{(h, d)}_{i, j},
\end{equation}
\begin{equation}
a_{T_k} = \frac{1}{|R|} \sum_{i \in R, j \in T_k} \bar a_{i, j},
\end{equation}
\noindent where $R$ denotes the set of response tokens, and $|\mathcal{D}|$ and $H$ represent the dataset size and the number of heads, respectively.

\subsubsection{Attention per Visual Token}
Given the average attention matrix across all heads and samples, denoted as $\bar a$ (Eq.~\ref{eq:avg_attention}), we define the average attention per visual token as follows:
\begin{equation}
a_V = \frac{1}{|V|} \sum_{0 \le i < N, j \in V} \bar a_{i, j},
\end{equation}
\noindent where $V$ is the set of visual tokens and $N$ is the number of tokens in the sequence.

\subsubsection{Attention Entropy}
Given the average attention matrix across all heads and samples, denoted as $\bar a$ (Eq.~\ref{eq:avg_attention}), we define the attention entropy as follows:
\begin{equation}
\mathcal{H} = - \frac{1}{N^2} \sum^{N}_{i=1} \sum^{N}_{j=1} \bar a_{i, j} log( \bar a_{i, j} + \epsilon),
\end{equation}
\noindent where $N$ is the number of tokens in the sequence, and $\epsilon = 10^{-8}$.

\end{document}